%% file: main.tex
\documentclass{svproc}
\usepackage{url}

\usepackage{cite}

\usepackage{siunitx}
\usepackage{graphicx} 
\usepackage{hyperref} 
\usepackage{amsmath}  
\usepackage{amsfonts} 
\usepackage{amssymb}  
\usepackage{comment}
\usepackage{multirow}
\usepackage[export]{adjustbox}
\usepackage{caption}
\usepackage{xcolor}

\newcommand{\rev}[1]{{#1}}

\definecolor{somegray}{rgb}{0.5, 0.5, 0.5}

\usepackage{tikz}
\usepackage{textcomp}
\usepackage[doipre={DOI:~}]{uri}
\usepackage{lipsum}
\newcommand\copyrighttext{%
  \footnotesize This paper has been accepted for publication in the ERF 2024 conference.
  }
\newcommand{\copyrightnotice}{%
\begin{tikzpicture}[remember picture,overlay]
\node[anchor=south,yshift=10pt] at (current page.south) {\fbox{\parbox{\dimexpr\textwidth-\fboxsep-\fboxrule\relax}{\copyrighttext}}};
\end{tikzpicture}%
}

\begin{document}
\mainmatter              
\title{Optimized Deployment of Deep Neural Networks for Visual Pose Estimation on Nano-drones}
\titlerunning{Optimized Deployment}  
%
\author{Matteo Risso\inst{1} \and Francesco Daghero\inst{2} \and Beatrice Alessandra Motetti\inst{1} \and Daniele Jahier Pagliari\inst{1} \and Enrico Macii\inst{2} \and Massimo Poncino\inst{1} \and Alessio Burrello\inst{2}}
\authorrunning{Matteo Risso et al.} 
\tocauthor{Matteo Risso, Francesco Daghero, Beatrice Alessandra Motetti, Daniele Jahier Pagliari, Enrico Macii, Massimo Poncino, Alessio Burrello}
\institute{
Department of Control and Computer Engineering,\\
\and
Interuniversity Department of Regional and Urban Studies and Planning,\\
Politecnico di Torino, Turin 10129, Italy,\\
\email{first\_name.first\_surname@polito.it}
}

\maketitle              
\copyrightnotice

\begin{abstract}
\input{sec/abstract}
\keywords{Nano-drones, TinyML, NAS, CNN, Fused Layers}
\end{abstract}
\section{Introduction and Related Works}
\input{sec/introduction}

\section{Materials and Methods}
\input{sec/method}
\section{Experimental Results}
\input{sec/results}
\section{Conclusions}
\input{sec/conclusions}

\bibliographystyle{spmpsci}
\bibliography{references}
%
%








\end{document}

%% file: sec/abstract.tex
%
%
%
\looseness=-1
Miniaturized autonomous unmanned aerial vehicles (UAVs) are gaining popularity due to their small size, enabling new tasks such as indoor navigation or people monitoring. Nonetheless, their size and simple electronics pose severe challenges in implementing advanced onboard intelligence.
This work proposes a new automatic optimization pipeline for visual pose estimation tasks using Deep Neural Networks (DNNs). The pipeline leverages two different Neural Architecture Search (NAS) algorithms to pursue a vast complexity-driven exploration in the DNNs' architectural space. The obtained networks are then deployed on an off-the-shelf nano-drone equipped with a parallel ultra-low power System-on-Chip leveraging a set of novel software kernels for the efficient fused execution of critical DNN layer sequences.
Our results improve the state-of-the-art reducing inference latency by up to 3.22$\times$ at iso-error.

%% file: sec/introduction.tex
Nano-sized unmanned aerial vehicles (UAVs), commonly named ``nano-drones'', are increasingly used for navigation in GPS-denied environments and to operate near humans, thanks to their small size (sub-\SI{10}{\centi\meter}) and low weight (sub-\SI{40}{\gram}). However, their limited on-board computational and memory capacity pose substantial challenges to achieving full autonomy. In particular, they make it impossible to utilize large deep learning models for perception~\cite{ai-deck,frontnet}.
\looseness=-1
A particularly relevant task for nano-UAVs is \textit{human pose estimation}~\cite{frontnet}, which enables applications such as “people monitoring” and “follow-me”. Recent research on this task has concentrated on optimizing tiny Convolutional Neural Networks (CNNs) to operate within tight nano-UAVs hardware constraints, marking a significant stride towards obtaining reasonable perceptual performance on such platforms~\cite{frontnet,cereda_deep_2023,date2024}. The work of~\cite{cereda_deep_2023}, in particular, underscored the critical role of automated optimization methods such as \textit{Neural Architecture Search} (NAS) in facilitating the design of efficient architectures that balance computational efficiency with task performance. However, \cite{cereda_deep_2023} only scratched the surface of a potentially vast research direction, applying a NAS algorithm aimed at shrinking an input CNN architecture (called ``seed'') through the elimination of unimportant feature maps~\cite{pitjournal}, closely resembling \textit{structured pruning}. Other NAS methods allow exploring broader (yet less fine-grained) search spaces, e.g., by selecting between different alternatives for each layer of the CNN~\cite{mnasnet,darts,plinio,duccio}. In this work, we show that the sequential application of these two families of NASs (layer selection and model shrinking) can yield superior results in terms of the pose estimation accuracy versus latency and memory trade-offs.

In addition to an optimized network architecture, another key component to enable real-time CNN-based perception on nano-UAVs is the availability of efficient low-level software kernels, fully exploiting the hardware available onboard. To this end, this work proposes the usage of optimized kernels for the execution of \textit{fused DepthWise (DW) and PointWise (PW) convolution} on the Parallel Ultra Low-Power (PULP) multi-core clusters available in recent nano-UAV platforms~\cite{ai-deck}. Sequences of DW and PW layers are common in tiny CNNs (inspired by MobileNets~\cite{mobilenets}), and fusing them significantly reduces the amount of intermediate memory transfers, thus improving end-to-end latency compared to single-layer kernels for the same hardware, such as the ones in~\cite{garofalo2019pulp}.

\looseness=-1
Through the combined optimization of a CNN architecture (with two chained NAS steps) and of the corresponding inference software stack, we outperform the state-of-the-art (SoTA) on human pose estimation for nano-UAV-class devices~\cite{cereda_deep_2023,date2024}, obtaining up to 13.78\% lower Mean Absolute Error (MAE), or reducing latency by up to 3.22$\times$ at iso-error. Our work highlights the key importance of multi-level automated deployment flows for tinyML on tiny drones.

%% file: sec/method.tex
\begin{figure}
    \centering
    \includegraphics[width=0.99\textwidth]{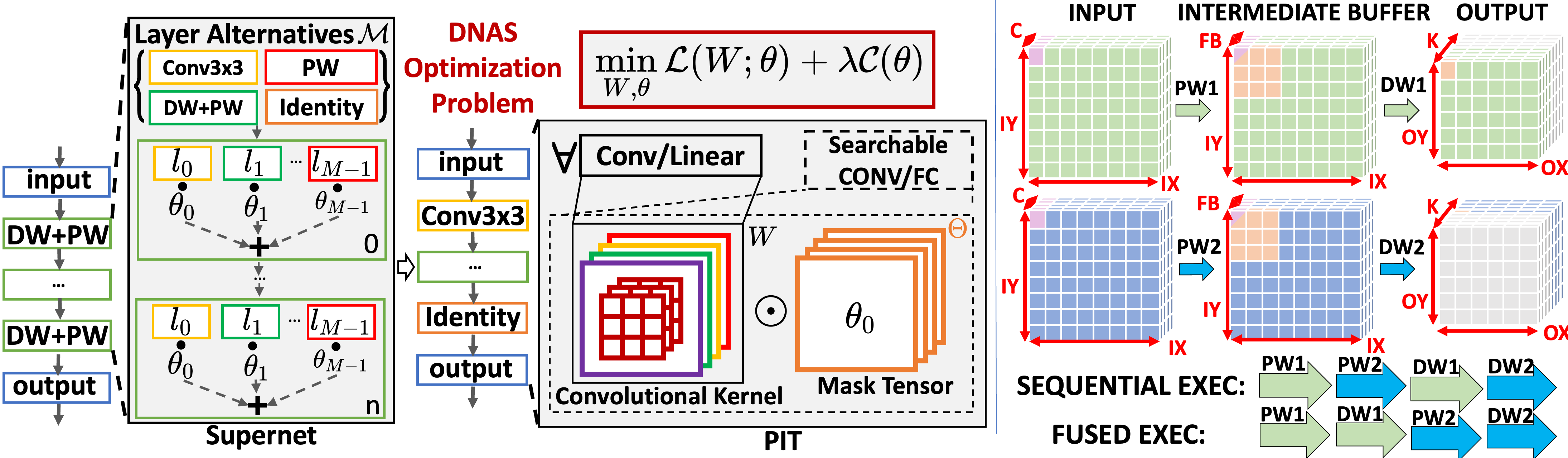}
    \vspace{-0.3cm}
    \caption{NAS-based optimization flow (left); Optimized PW+DW kernel (right).}
    \label{fig:opt-flow}
    \vspace{-0.6cm}
\end{figure}

\paragraph{Target Platform:}~\label{subsec:platform}
We focus on the Bitcraze Crazyflie 2.1 nanodrone equipped with a greyscale camera and the GAP8 SoC~\cite{gap8}. GAP8 comprises a single-core fabric controller (FC) and an 8-core PULP cluster (CL). The FC orchestrates memory transfers and delegates demanding computations to the CL. CL cores share a 64 kB L1 memory and the SoC includes a 512 kB L2 memory. A Direct Memory Access (DMA) unit handles transfers between the two memories.
%
%
\paragraph{Complexity-driven Architecture Search: }~\label{subsec:nas}
Fig.~\ref{fig:opt-flow} (left) shows the proposed architecture optimization flow, which combines two SotA NASs, \textit{Supernet}~\cite{darts} and \textit{PIT}~\cite{pitjournal}, using the implementations of the PLiNIO open-source library~\cite{plinio}.
Both are so-called Differentiable NASs (DNASs), i.e., they jointly train (with gradient descent) the standard weights of the network $W$ and some additional parameters $\theta$, which control architectural choices such as the type of layer (for \textit{Supernet}) or the number of output features of each layer (for \textit{PIT}).
The red box of Fig.~\ref{fig:opt-flow} shows the loss function minimized by both NASs, where $\mathcal{L}$ is the task-specific loss (e.g., Mean Squared Error) and $\mathcal{C}$ is an added complexity term such as the number of parameters or operations of the network, as a function of $\theta$. The balance between the two is controlled by the scalar $\lambda$. \rev{In particular, we used the size complexity defined in~\cite{pitjournal}.}
%
%
%
%

We use a MobileNetV1 architecture as blueprint for the SuperNet step, since it demonstrated SotA performance on human-to-nanodrone pose estimation~\cite{cereda_deep_2023, date2024}. For each DW+PW block of the original network, the NAS selects between: i) The original block; ii) A single PW layer; iii) A standard 2DConv with 3$\times$3 filter; iv) A ``no-operation'' to optionally skip the block.
As depicted in Fig.~\ref{fig:opt-flow}, each of the $\mathcal{M}=4$ alternatives is coupled with a $\theta_{i}$ parameter. At the end of training, the alternative associated with the largest $\theta_{i}$ is selected.
%
%
%
The architectures obtained with the Supernet are further optimized with PIT, which applies a fine-grained structured pruning, eliminating unimportant output channels from each layer.
Namely, the weight tensor of each Conv or linear layer is paired with a set of binary trainable masks $\theta$, which are trained to control whether a specific feature is removed ($\theta_{i} = 0$) or kept ($\theta_{i} = 1$).
\vspace{-0.35cm}
\paragraph{Fused Kernel for efficient inference:} 
DW layers are notoriously difficult to accelerate due to their more limited data reuse options compared to standard 2D Conv. PULP-NN~\cite{garofalo2019pulp}, a SoTA open-source kernel library for GAP8, handles this by changing the input data layout from Height-Width-Channels (HWC) to Channels-Height-Width (CHW). However, this causes an increase in data transfers, as the re-ordering operation has to be repeated multiple times when tiling the layers (i.e., loading parts of the data in L1 memory before performing computations)~\cite{burrello2020dory}.
%
%
To reduce this overhead, we implement a new fused kernel, computing a PW+DW sequence entirely in L1, as shown in Fig.~\ref{fig:opt-flow} (right).
%
%
We accelerate PW+DW (and not DW+PW) sequences, as this enables exploiting the independence of DW computations across input/output channels (C/K). Specifically, we compute a subset (FB) of the PW output channels storing them in an additional L1 buffer of size IX*IY*FB (IX/IY = input rows/columns); then, we execute the DW operation on such buffer. The whole process is repeated until all K output channels have been produced.
Since the DW is parallelized over channels, we set $FB=8$, i.e., equal to the cores of GAP8, to maximize utilization with the
%
%
minimum possible L1 memory overhead.

%% file: sec/results.tex
\begin{figure}[htbp]
    \centering
    \begin{minipage}[c]{0.44\textwidth}
        \includegraphics[width=.9\textwidth]{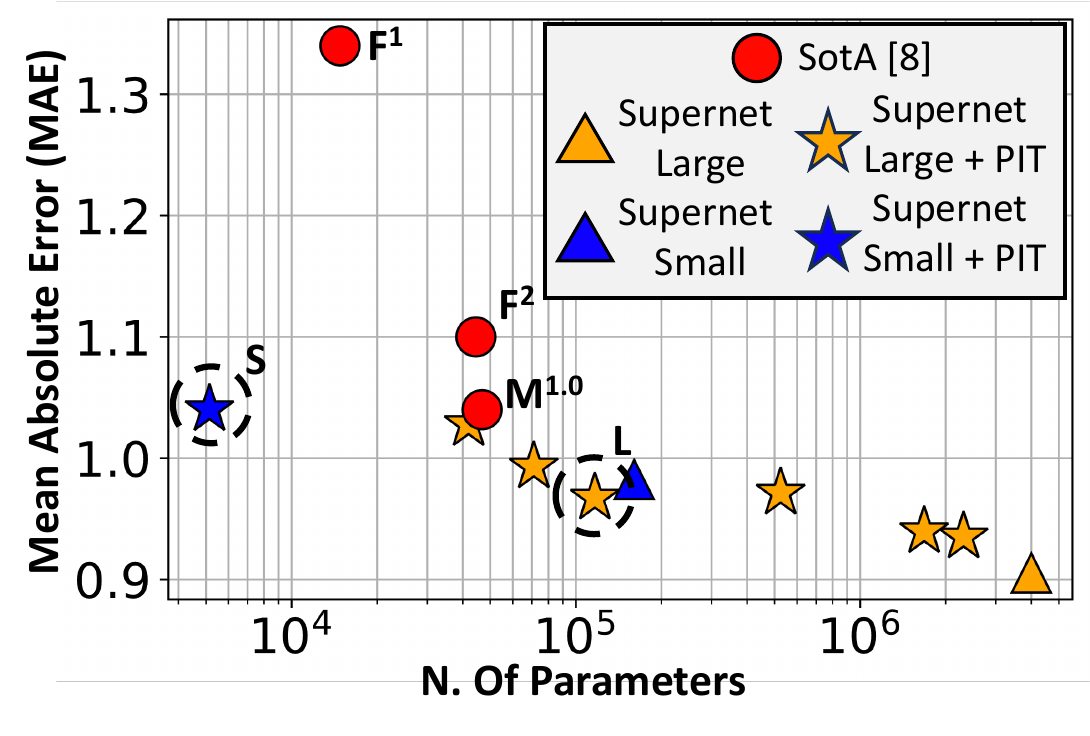}
        \vspace{-0.5cm}
        \caption{Optimized architectures vs SotA from~\cite{date2024}.}
        \label{fig:pareto}
    \end{minipage}
    \hfill
    \begin{minipage}[c]{0.4\textwidth}
        \centering
        {\scriptsize
        \begin{tabular}{l|l|l|l|l}
        \textbf{Model} & \textbf{Ker} & \multicolumn{1}{l}{\textbf{MAE}} & \multicolumn{1}{|l|}{\textbf{Lat{[}ms{]}}} & \multicolumn{1}{|l|}{\textbf{Mem{[}kB{]}}} \\ \hline\hline
        \multicolumn{5}{l}{\textbf{SoTA}}                                                  \\ \hline
        $F^1$               & U & 1.34                  & 7.06  & 14.8                     \\
        $F^2$               & U & 1.1                   & 8.82  & 44.5                     \\
        $M^1$               & U & 1.04                  & 21.76 & 46.8                     \\\hline
        \multicolumn{5}{l}{\textbf{Ours}}                                                  \\ \hline
        \multirow{2}{*}{S} & U & \multirow{2}{*}{1.04} & 7.4   & \multirow{2}{*}{5.14}   \\
                            & F &                       & 6.76  &                         \\
        \multirow{2}{*}{L} & U & \multirow{2}{*}{0.97} & 32.5  & \multirow{2}{*}{116.63} \\
                            & F &                       & 31.43 &                         \\\hline
        \end{tabular}}
        \captionof{table}{Deployment results.}
        \label{tab:deployment}
    \end{minipage}
    \vspace{-0.7cm}
\end{figure}

\looseness=-1
We train and test our CNNs on the dataset introduced in~\cite{frontnet}, and with the same data splits of~\cite{date2024}
%
Fig.~\ref{fig:pareto} shows the results obtained with the cascaded application of the Supernet and PIT NASs compared with the SotA networks (red circles) of~\cite{date2024} in the N. of Parameters vs. MAE plane. All the results are uniformly quantized to INT8 data format using~\cite{plinio}.
We use a slightly modified version of the test set, as in~\cite{date2024}, where the labels have been adjusted to compensate for a calibration inaccuracy of the measurement tools detected through manual inspection.

\looseness=-1
We apply the SuperNet NAS using two different MobileNetV1 variants as blueprints: a standard one (denoted as \textit{Large}) and one with a width-multiplier of $0.25\times$ (\textit{Small}).
The orange and blue triangles of Fig.~\ref{fig:pareto} denote two promising output architectures obtained with this first NAS step.
In particular, we selected the blue triangle (SuperNet Small) as the smallest network achieving a MAE $<$ 1. Conversely, we selected the orange one (SuperNet Large) as the network achieving the lowest MAE.
The found architectures details are as follows: SuperNet Large substitutes the first three DW+PW layers of the vanilla MobileNetV1 with standard 2DConv layers.
Conversely, SuperNet Small substitutes only the first DW+PW block with a 2DConv, while skipping entirely the last DW+PW block.

We then apply PIT to both these models, obtaining the rich collection of Pareto-optimal architectures, depicted in Fig.~\ref{fig:pareto} with blue and orange stars.
Noteworthy, our smallest model achieves a $9\times$ size compression at iso-MAE w.r.t. the most accurate SotA model~\cite{date2024}.
Conversely, in the large N. of Parameters regime, our solutions can improve the SotA by up to 13.78\%.
\rev{Note that these results showcase the benefits of the combined usage of both NASs, given that SotA networks have been obtained using PIT only~\cite{cereda_deep_2023}.}
\vspace{-0.5cm}
%

We deploy models on GAP8 mirroring the setup of~\cite{date2024}.
Table~\ref{tab:deployment} shows the deployment results of selected CNNs from Fig.~\ref{fig:pareto} (those labeled with a letter) from our work and the SoTA. We deploy the smallest (S) and most accurate (L) CNNs found by our NAS chain, excluding models that do not fit the L2 memory of GAP8 (512 kB). 
For each model, we report the test set MAE, the weight memory footprint (Mem), and the latency (Lat). We also report the kernel used for all PW+DW sequences (Ker), which is either our proposed fused implementation (F) or an unfused one (U) from vanilla PULP-NN.

When considering the S model, we achieve the same MAE at far lower latency (-65\%) w.r.t. the most accurate SoTA model (M1).
The speedup increases to 68.1\% (a further 8.6\% reduction) when utilizing our fused PW+DW kernel.
If we compare the same model to the least accurate SoTA one (F1), we achieve significantly better MAE (-0.3) with more than 30$\times$ fewer parameters. When using fused kernels, we also still reduce latency by 4.2\%.
%
The L model, on the other hand, outperforms the most accurate SoTA CNN in terms of MAE (6.98\% reduction).
Fused kernels are less beneficial for this model, as most of the latency is due to the initial standard convolutional layers.
Nonetheless, they still grant a latency reduction of 3.27\% compared to a standard deployment.
Noteworthy, as shown in~\cite{cereda_deep_2023}, reducing perception latency (i.e., improving the frames per second that the model can process) has been shown very beneficial to improve the performance of nano-drones control loops.

%% file: sec/conclusions.tex
We have presented a multi-stage, fully-automated optimization pipeline for visual human-to-nanodrone pose estimation, including two chained NAS methods, respectively for layer selection and model pruning, and an optimized implementation of fused PW and DW convolutions to improve CNN inference latency, and consequently the drone's controller reaction time, by cutting intermediate memory transfers. Overall, our work serves to demonstrate the fundamental importance of deployment optimization pipelines for TinyML on tiny drones.